\documentclass[conference]{IEEEtran}
\IEEEoverridecommandlockouts
\usepackage{cite}
\usepackage{amsmath,amssymb,amsfonts}
\usepackage{algorithmic}
\usepackage{graphicx}
\usepackage{textcomp}
\usepackage{subfiles}
\usepackage{subfigure}
\usepackage{booktabs}
\usepackage{amsmath}
\usepackage{multirow}
\usepackage{xcolor}
\usepackage{algorithm}
\usepackage{subfiles}
\def\BibTeX{{\rm B\kern-.05em{\sc i\kern-.025em b}\kern-.08em
    T\kern-.1667em\lower.7ex\hbox{E}\kern-.125emX}}
\begin{document}

\title{Towards Better Understanding with Uniformity and Explicit Regularization of Embeddings in Embedding-based Neural Topic Models\\
}
\makeatletter
\newcommand{\linebreakand}{%
  \end{@IEEEauthorhalign}
  \hfill\mbox{}\par
  \mbox{}\hfill\begin{@IEEEauthorhalign}
}
\makeatother

\author{\IEEEauthorblockN{1\textsuperscript{st} Wei Shao}
\IEEEauthorblockA{\textit{Department
 of Computer Science} \\
\textit{City University of Hong Kong}\\
Kowloon Tong, Hong Kong SAR \\
weishao4-c@my.cityu.edu.hk}
\and
\IEEEauthorblockN{2\textsuperscript{nd}Lei Huang}
\IEEEauthorblockA{\textit{Department
 of Computer Science} \\
\textit{City University of Hong Kong }\\
Kowloon Tong, Hong Kong SAR \\
lhuang-93@my.cityu.edu.hk}
\and
\IEEEauthorblockN{3\textsuperscript{rd} Shuqi Liu}
\IEEEauthorblockA{\textit{Department
 of Computer Science} \\
\textit{City University of Hong Kong}\\
Kowloon Tong, Hong Kong SAR \\
shuqiliu4-c@my.cityu.edu.hk}
\linebreakand 
\IEEEauthorblockN{4\textsuperscript{th} Shihua Ma}
\IEEEauthorblockA{\textit{Department
 of Mechanical Engineering} \\
\textit{City University of Hong Kong}\\
Kowloon Tong, Hong Kong SAR \\
shihuama-2@my.cityu.edu.hk}
\and
\IEEEauthorblockN{5\textsuperscript{th} Linqi Song$^*$}
\IEEEauthorblockA{\textit{Department
 of Computer Science} \\
\textit{City University of Hong Kong}\\
Kowloon Tong, Hong Kong SAR \\
\textit{City University of Hong Kong}\\
\textit{Shenzhen Research Institute}\\
Shenzhen, China\\
linqi.song@cityu.edu.hk}
\thanks{$^*$Corresponding author.}
}
\maketitle

\begin{abstract}
Embedding-based neural topic models could explicitly represent words and topics by embedding them to a homogeneous feature space, which shows higher interpretability. However, there are no explicit constraints for the training of embeddings, leading to a larger optimization space. Also, a clear description of the changes in embeddings and the impact on model performance is still lacking. In this paper, we propose an embedding regularized neural topic model, which applies the specially designed training constraints on word embedding and topic embedding to reduce the optimization space of parameters. To reveal the changes and roles of embeddings, we introduce \textbf{uniformity} into the embedding-based neural topic model as the evaluation metric of embedding space. On this basis, we describe how embeddings tend to change during training via the changes in the uniformity of embeddings. Furthermore, we demonstrate the impact of changes in embeddings in embedding-based neural topic models through ablation studies. The results of experiments on two mainstream datasets indicate that our model significantly outperforms baseline models in terms of the harmony between topic quality and document modeling. This work is the first attempt to exploit uniformity to explore changes in embeddings of embedding-based neural topic models and their impact on model performance to the best of our knowledge.
\end{abstract}


\section{Introduction}
The topic model could model the relation between words and documents via the latent topics. In detail, a topic model could discover which topics are more frequent in the given documents and which words are more frequent in the given topics. The above information could be represented by document-topic distribution and topic-word distribution. Owing to its ability to discover this kind of latent structure of documents, topic model are widely applied in sentiment analysis\cite{xiong2018short}, document classification\cite{Seifollahi2021AnET}, key-phrase extraction\cite{li2013novel} and other tasks.

Before the rise of neural networks, probability graph topic models\cite{Blei2012ProbabilisticTM, hofmann1999probabilistic} such as LDA\cite{blei2003latent} were popular. In the age of neural networks, the neural topic model has become the mainstream model. Several kinds of neural topic models, including VAE-based neural topic models, adversarial topic models, and embedding-based neural topic models, were proposed and displayed better performance on the topic quality and document modeling. In these topic models, embedding-based topic models could explicitly represent words and topics by embedding them to a vector space, which shows higher interpretability. A typical embedding-based topic model is ETM, proposed by \cite{dieng2020topic}. It has a simple architecture and approach to training and achieves a better performance than \cite{blei2003latent} and the \cite{Miao2016NeuralVI}, which are typical probability graph topic models and VAE-based topic models, respectively.

Although embedding-based neural topic model has satisfactory performance and better interpretability, it utilizes the same loss function as the VAE-based neural topic model, which do not pay attention to the training constraints of topic embedding and word embedding. This means that the embeddings will optimized in a large parameter space and thus increase difficulty of optimization. 

To further tackle this problem, we propose an \textbf{E}mbedding \textbf{R}egularization \textbf{N}eural \textbf{T}opic \textbf{M}odel (ERNTM) introducing specially designed constraints into the training process of embeddings. In addition to reducing the search space, we also hope that these constraints can make embeddings reflect certain characteristics. In detail, for topic embedding, a given topic is expected to differ from other topics for a better diversity. To this end, we added a regularization term which is relevant to the similarity between topic embeddings to the loss function. Then, we expect that word embedding could obtain some frequency information from documents. Therefore, we add a regularization term representing the distance between word embedding and projection of documents in the same space.. We conduct a series of experiments to show our model's better performance over other strong embedding-based topic models and neural topic models.

Another critical problem is the lack of in-depth exploration of embedding variation and its impact on model performance, which is essential to understanding this kind of model. None of the works attempt to describe the changes of embedding during training quantitatively, so it is not clear what is a good embedding in an embedding-based topic model and how the embedding affects the model's performance. To narrow this gap, in this paper, we introduce uniformity\cite{Wang2020UnderstandingCR} as a tool to perform a quantitative description of the embedding, which could measure how well the embeddings are uniformly distributed. The lower uniformity value implied the higher diversity of embeddings. On this basis, we could show the changes in embeddings during training by their uniformity changes. Then, we present how these embeddings' changes influence the neural topic model via ablation study, which reveals the internal mechanism of the embedding-based topic model from the view of embedding's uniformity.

To summarize, the main contributions of this paper are as follows:
\begin{itemize}
    \item We propose an \textbf{E}mbedding \textbf{R}egularized \textbf{N}eural \textbf{T}opic \textbf{M}odel (ERNTM), which applies the training constraint on the word embedding and the topic embedding to reduce their parameter space for better optimization.
    \item We conduct a series of comparison experiments on two mainstream datasets and the results indicate that our model significantly outperforms baseline models in terms of the harmony between the topic quality and document modeling.
    \item Based on uniformity, we quantitatively present how the word embedding and topic embedding change in the training process and show how they influence the topic model's performance via ablation study.
\end{itemize}

To our best knowledge, we are the first to introduce uniformity into embedding-based topic models to evaluate the topic and word embedding and provide a novel view to explain the embedding-based neural topic model.

\section{Related Work}

With the development of deep learning, neural networks have been widely applied to topic modeling, achieving better performances than traditional topic models. Generally, we can divide these topic models into three classes. 

First is the VAE-based neural topic model. These models are based on variational auto-encoder architecture\cite{Kingma2014AutoEncodingVB} and utilize variational inference to obtain the document-topic distribution. NVDM\cite{Miao2016NeuralVI} is a typical model, which assumes that the document-topic distribution conforms to the Gaussian distribution and obtains document-topic distribution by variational methods and sampling. Then a multi-layer perceptron is used to transform this distribution to a probability proportion over vocabulary. The weight of this multi-layers perceptron could be converted to the topic-word distribution. Later, some people improved NVDM and proposed GSM\cite{miao2017discovering}, which normalized the distribution over topics to train the model more stably and rapidly. Then someone proposed ProdLDA\cite{Srivastava2017AutoencodingVI} on this basis, replacing the Gaussian distribution with the Dirichlet distribution as the prior distribution. Most VAE-based models optimize the variational distribution with KL divergence distance. However, there are some VAE-based topic models utilizing different distance functions when optimizing the variational distribution. A classic work is \cite{Nan2019TopicMW}, which uses the Wasserstein distance to regulate the variational distribution to alleviate the KL collapse problem.

Another is the adversarial topic model. \cite{Wang2019ATMAT} proposed ATM, which first uses adversarial training for topic modeling. It achieves a better performance on topic coherence and some downstream tasks. After that, \cite{Wang2020NeuralTM} presents a neural topic model with bidirectional adversarial training. However, these two adversarial topi models are unable to infer topic distributions for given documents or utilize available document labels. To overcome such limitations, \cite{Hu2020NeuralTM} introduced the cycle-consistent adversarial training into the neural topic model.

The third is the embedding-based neural topic model. NTM\cite{ding-etal-2018-coherence} and its variants utilize word embedding and the weight of a multi-layer perceptron to construct topic embedding and produce the topic-word distribution.
ETM\cite{dieng2020topic} explicitly embeds words and topics in the same semantic space and represents topic-word distribution as the cross-product of two embedding matrices. Due to the full consideration of document information and semantic information of words, the effect of ETM is superior to some other neural topic models and traditional topic models.

However, embedding-based neural topic models suffer from two problems: the embedding's optimization space is ample and the relationship between the embeddings and the model's performance is still unclear. To solve the above problems, we proposed an embedding regularization neural topic model applying training constraints on embeddings to narrow their search space in training. This model shows higher performance on the topic quality and document modeling than other baselines. Also, we introduce uniformity to explain the relationship between embeddings and model performance and conduct a series of experiments to present how the word embedding and topic embedding change in the training process and how these embeddings' influence the model's performance.

\section{Backgroud}
\subsection{Problem Definition}
Given a corpus with $M$ documents $\{\boldsymbol{x_1, x_2, ..., x_M}\}$ whose vocabulary $\{w_1, w_2, ..., w_V\}$ contains $V$ different words and each document $\boldsymbol{x_i} \in R^{V}$ is represented by bag-of-words model, topic models aims to model these documents via $K$ latent topics. There are several assumptions in topic modeling. First, a document $\boldsymbol{x_i}$ should include several topics, which means each document has a distribution $\boldsymbol{\theta_i} \in R^{1 \times K}$ over the latent topics $\boldsymbol{Z}=\{z_1, z_2, ..., z_K\}$. Second, the proportions over words $\boldsymbol{\beta} \in R^{K \times V}$ for different topics are different. Third, we do not know the exact meaning of each topic because they are latent variables. This also hinders the interpretability of the topic models. According to the conditions above, for a document, we can estimate the probability of each word in this document via $(\boldsymbol{\theta_d} \times \boldsymbol{\beta}) \in R^{1 \times V}$.

The modeling process could be optimized by maximizing the likelihood of words(the negative likelihood is loss function) in documents:
\begin{equation}
    L = \sum^{M}_{i=1}logp(\boldsymbol{x_i})\,, 
\end{equation}
\begin{equation}
    p(\boldsymbol{x_i}) = \prod^{V}_{j=1} (\sum^{K}_{k=1}p(z_k|\boldsymbol{x_i})p(w_j|z_k))^{x_{ij}}\,,
\end{equation}
\begin{equation}
    p(\boldsymbol{x_i}) = \prod^{V}_{j=1} (\boldsymbol{\theta_{i}} \times \boldsymbol{\beta})^{x_{ij}}\,.
\end{equation}

Therefore, the problem is transformed to how to calculate the document-topic distribution $\boldsymbol{\theta}$ and the topic-word distribution $\boldsymbol{\beta}$.

\subsection{Embedding-based Topic Models}
Embedding-based topic models embed words and topics into a vector space, which produces two embedding matrixes: word embedding matrix $\boldsymbol{\rho} \in R^{V \times E}$ and topic embedding matrix $\boldsymbol{\alpha} \in R^{K \times E}$. The topic-word distribution $\boldsymbol{\beta} = softmax(\boldsymbol{\alpha} \times \boldsymbol{\rho})$ means that words closer to a certain topic are assigned higher probability in the topic-word distribution. We can visualize embeddings into a 2/3-dimension space with dimension reduction methods, which provides an explainable view for the topic modeling. However, this view only provides an intuitive explanation for results, without a quantitative explanation for the role of embedding in the topic model, which is our focus in this work.

\section{Method}
\begin{figure*}[tp]
    \centering
    \includegraphics[width=\textwidth]{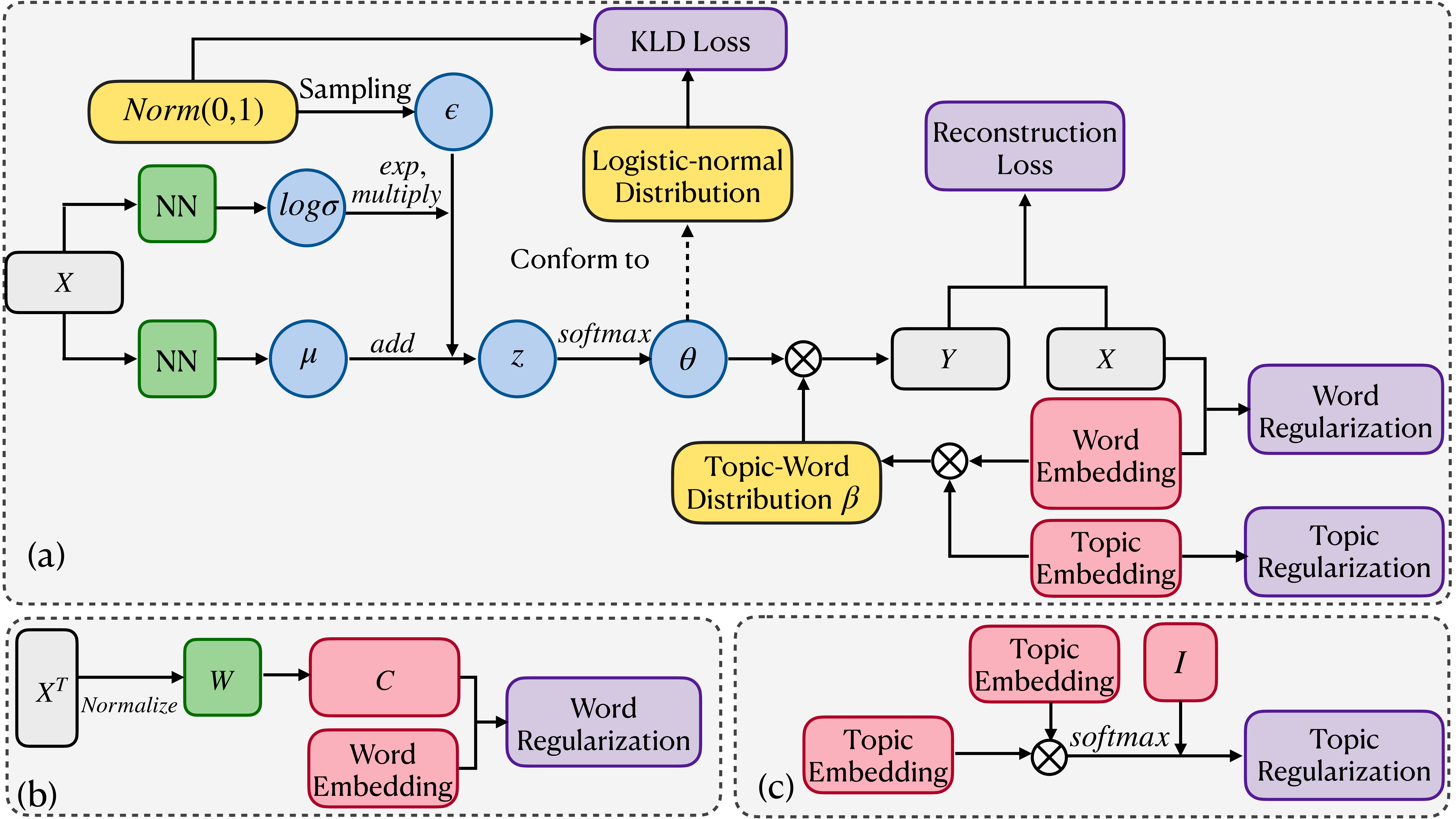}
    \caption{(a) shows our model's architecture based on batch training framework. This framework will produce four loss terms and our model is optimized by minimizing the weighted sum of them simultaneously. (b) indicates the computation process of word regularization. (c) presents how to compute the topic regularization.}.
    \label{fig:model}
\end{figure*}

\subsection{Overview}
This section will introduce our proposed model: an embedding regularized neural topic model. This model takes the bag-of-words representation of documents as input. 
For the modeling process, it first computes the document-topic distribution $\boldsymbol{\theta}$ via variational inference\cite{Kingma2014AutoEncodingVB}. Next, the model will calculate the topic-word distribution $\boldsymbol{\beta}$, which is represented as the matrix dot product of the topic embedding matrix and the word embedding matrix. After obtaining the document-topic distribution $\boldsymbol{\theta}$ and the topic-word distribution $\boldsymbol{\beta}$, we can model the likelihood of the documents with the formulations mentioned in the background. The detailed process is shown in the Fig.~\ref{fig:model} and we will present a more detailed description in the following parts. Before that, we need to denote our notion and variable representation.

\subsection{Notion}
Consider a document set $\boldsymbol{D} = \{d_1, d_2, ...,d_M\}$  and its bag-of-words representation $\{\boldsymbol{x_1}, \boldsymbol{x_2}, ..., \boldsymbol{x_M}\}$ and a vocabulary $\boldsymbol{W} = \{w_1, w_2,...,w_V\}$ formed by words in documents.
Each word in $\boldsymbol{W}$ is embedded to $E$-dimension Hilbert spaces. We use $\boldsymbol{\rho} \in R^{V \times E}$ to represent the word embedding matrix. Similarly, we embed the latent topic set $\boldsymbol{Z} =\{z_1, z_2, ...,z_K\}$ into the same $E$-dimensional space. We use $\boldsymbol{\alpha} \in R^{K \times E}$ to represent the topic embedding matrix.
In our model, each document is in the form of bag-of-words. 
The neural networks model the probabilities over topics (e.g., $\boldsymbol{\theta_i} \in R^{K}$) conditioned on document $d_i$ with the documents' embedding vectors $\boldsymbol{x_i} \in R^{V}$ as input. The matrix $\boldsymbol{\beta} \in R^{K\times V}$ denotes the distribution over vocabulary $\boldsymbol{W}$ whose $k$-th row represents that of topic indexed by $k$.

\subsection{Modeling Process}
As mentioned before, the modeling process could be transformed to the computation of the $\boldsymbol{\theta}$ and $\boldsymbol{\beta}$. The computation process is shown in Fig.~\ref{fig:model} (a). We will describe how to obtain the document-topic distribution and the topic-word distribution, respectively, in the following parts.

\subsubsection{Document-Topic Distribution $\boldsymbol{\theta}$}
In Fig.~\ref{fig:model} (a),  $\boldsymbol{X} \in  R^{B \times V}$ is a batch of document vectors, where $B$ is the number of documents in a batch and $\boldsymbol{\theta} = \{\boldsymbol{\theta_1}, \boldsymbol{\theta_2}, ..., \boldsymbol{\theta_B} \}$. In our model, $B$ is set to $1000$.

Being agnostic of the real document-topic distribution, researchers usually make assumptions that $\boldsymbol{\theta_d}$ ($\theta_{d,k} = p(z_k|d)$) conforms to the logistical-normal distribution and reparameterize its parameters with neural networks. 

Concretely, $\boldsymbol{X}$ is taken as input by three fully connected layers followed by two Tanh activation functions. The neural networks output two vectors: $\log\boldsymbol{\sigma} \in R^{B \times K}$ and $\boldsymbol{\mu} \in R^{B \times K}$. Then a variable $\boldsymbol{\epsilon}\in R^{B \times K}$ is obtained by sampling from a standard normal distribution $\boldsymbol{N(0,I)}$. $\boldsymbol{\epsilon}$ can produce $\boldsymbol{z} \in R^{B \times K}$ by $\boldsymbol{\epsilon} \odot \boldsymbol{\sigma} + \boldsymbol{\mu}$. After that, the document-topic distribution $\boldsymbol{\theta} \in R^{B \times K}$ is obtained by softmax nonlinearities. This process could be formulated as:

\begin{equation}
   \boldsymbol{\mu} = NN(\boldsymbol{X};\boldsymbol{v_{\mu}})\,,
\end{equation}
\begin{equation}
    \boldsymbol{\sigma} = NN(\boldsymbol{X};\boldsymbol{v_{\sigma}})\,,
\end{equation}
\begin{equation}
    \boldsymbol{z} = \boldsymbol{\epsilon} \odot \boldsymbol{\sigma} + \boldsymbol{\mu}, \boldsymbol{\epsilon} \sim \boldsymbol{N(0, I)}\\,
\end{equation}
\begin{equation}
    p(\boldsymbol{Z}|\boldsymbol{X}) = \boldsymbol{\theta} = softmax(\boldsymbol{z})\,,
\end{equation}
where the $\boldsymbol{v_{\mu}}$ and $\boldsymbol{v_{\sigma}}$ are the parameters of the neural networks.
This distribution is the variational distribution we construct to fit the real document-topic distribution. Therefore we can approximate the true distribution with $\theta$ by reducing their KL divergence in the training process. (KL divergence between two distributions) which is represented as:
\begin{equation}
    L_{KLD} = -\frac{\sum_{i}^{B}\sum_{j}^{K}(1 + \log\boldsymbol{\sigma}_{i,j} - \boldsymbol{\mu}_{i,j}^{2} - e^{\log\boldsymbol{\sigma}_{i,j}})}{B}\,.
\end{equation}

\subsubsection{Topic-Word Distribution $\boldsymbol{\beta}$}
With Bayesian marginalization, we can obtain the topic-word distribution $\boldsymbol{\beta} \in R^{K \times V}$ by
\begin{equation}
    p(\boldsymbol{W}|\boldsymbol{Z}) = \boldsymbol{\beta} = \boldsymbol{\alpha} \times \boldsymbol{\rho}^\top\,.
\end{equation}

Each element in vector $\boldsymbol{\beta}_{k}$ indicates corresponding word's probability to topic $k$.(e.g. $\boldsymbol{\beta}_{i,j} = p(w_j|z_k)$).

\subsubsection{Likelihood of the Document}
After obtaining document-topic distributions $\boldsymbol{\theta}$ and topic-word distribution $\boldsymbol{\beta}$, we can produce each word $w_i$'s probability in document $d$:
\begin{equation}
    \boldsymbol{Y}_{d,i} =\sum_{z_k}^{\boldsymbol{Z}}\boldsymbol{\theta}_{dk}\boldsymbol{\beta}_{k,i}\,,
\end{equation}
forming $\boldsymbol{Y} \in R^{B \times V}$. Then, we can calculate the marginal likelihood of documents and use the negative value of it as the standard reconstruction loss:
\begin{equation}
   L_{recons} =  \frac{-\sum_{i}^{B}\sum_{j}^{V}(\log(\boldsymbol{Y}_{i,j}) \odot \boldsymbol{X}_{i,j})}{B}\,.
\end{equation}

\subsection{Constraints on Embedding}
To reduce the optimization space of the embedding, we place an explicit constraint on the topic embedding and the word embedding, respectively, in the training process. We will show how constraints are implemented in the following parts.

\subsubsection{Constraint on Word Embedding}
To diminish the search space of parameters and make word embedding learn more frequency information from documents, we designed a loss term as the constraint, which is presented in Fig.~\ref{fig:model} (b).
We normalize the transpose of $\boldsymbol{X}\in R^{B\times V}$ and use the neural networks (two fully connected layers with LeakyReLU activation function and a drop layer) represented by $W$ to project the normalization result $\boldsymbol{X}^T \in R^{V\times B}$ to a matrix $\boldsymbol{C}\in R^{V\times E}$. Then, the loss term could be computed as:
\begin{equation}
    R_{word} = \frac{-\sum_{i}^{V}\sum_{j}^{E}2\log(\boldsymbol{\rho}_{i,j}-\boldsymbol{C}_{i,j})}{V \times E}\,.
\end{equation}

$\boldsymbol{X}^T \in R^{V\times B}$ represents a view of words instead of documents. The normalization of $\boldsymbol{X}^T$ contains frequency information of each word on all documents in current batch. And this information could be regarded as a type of representation for words. Then the network $W$ projects them into the same space as the word embedding. The loss term means the difference between the word embedding and the projection. By introducing this term to the final loss function, we expect to reduce this distance and thus the word embedding could learn more frequency information from the document representations.

\subsubsection{Constraint on Topic Embedding}
As shown in Fig.~\ref{fig:model} (c), to reduce the optimization space of the topic embedding and maximize the distance between topic embeddings ( e.g. scaling down the inner-product between any two different topic embeddings ), we designed a regularization term and introduced it into loss function as the topic embedding constraint, which is formalized as:
\begin{equation}
    R_{topic} = \frac{-\sum_{i}^{K}\!\sum_{j}^{K}\boldsymbol{I}_{i,j} \odot \log \boldsymbol{D}_{i,j}}{K}\,,
\end{equation}
\begin{equation}
    \boldsymbol{D} = softmax(\boldsymbol{\alpha} \times \boldsymbol{\alpha}^\top)\,.
\end{equation}
where $\boldsymbol{I}$ is the identity matrix. 

$softmax(\boldsymbol{\alpha} \times \boldsymbol{\alpha}^\top)$ could be regarded as the similarity of topics represented by the inner-product between topic embeddings. $R_{topic}$ is the cross-entropy between the similarity and the unit diagonal matrix. By reducing this loss term in the training process, we can lower the similarity between different topics (normalized inner-product $\boldsymbol{D}_{i,j, i!=j}$ approaches  zero). We hope this constraint could enhance the topic diversity of our model.


\subsection{Final Loss}
The final loss function is the combination of the above four terms:
\begin{equation}
    L_{total} = L_{KLD} + \lambda_1 L_{Recons} + \lambda_2 R_{topic} + \lambda_3 R_{word}\,.
\end{equation}
where $\lambda_i$ are the hyperparameters controlling the relative importance of regularization terms. In our experiments, the $\lambda_1 = \lambda_2 = \lambda_3$ = 1. In our experiment, we utilize the Adam\cite{Kingma2015AdamAM} optimizer to perform the gradient descent for the update of parameters in the training process.

\subsection{Algorithm}
Here, we use an algorithm flowchart to clearly present the whole process of our method.
\begin{algorithm} 
	\caption{Topic Modeling with ERNTM} 
	\label{alg3} 
	\begin{algorithmic}
		\REQUIRE Topic Embedding $\boldsymbol{\alpha}$, Word Embedding $\boldsymbol{\rho}$, variational inference network, $\boldsymbol{v_{\mu}}$ and $\boldsymbol{v_{\sigma}}$, network $W$
		\STATE Initialize the model and variational parameters
		\FOR{iteration i=1, 2, ...}
		\STATE Obtain a minibatch $\boldsymbol{X}$ of documents
		\STATE Compute $\boldsymbol{\beta_k} = softmax(\boldsymbol{\rho}^T \boldsymbol{\alpha}_k)$
		\FOR{each document $d$ in $\boldsymbol{X}$}
		    \STATE Normalize the bag-of-words representation $\boldsymbol{X_d}$
		    \STATE Compute $\boldsymbol{\mu_d}$=NN($\boldsymbol{X_d}$;$\boldsymbol{v_{\mu}}$)
		    \STATE Compute $\boldsymbol{\sigma_d}$=NN($\boldsymbol{X_d}$;$\boldsymbol{v_{\sigma}}$)
		    \STATE Sample $d$'s distribution over topics $\boldsymbol{\theta_d}$ $\sim$ $LN$($\boldsymbol{\mu_d}$, $\boldsymbol{\sigma_d}$)
		    \FOR{each word $w_{dn}$ in $d$}
		        \STATE Compute $p(w_{dn}|\boldsymbol{\theta_d})=\boldsymbol{\theta_{d}}^{T} \boldsymbol{\beta}_{.,w_{dn}}$
		    \ENDFOR
		\ENDFOR 
		\STATE Compute the KLD loss with variational inference $L_{KLD}$
		\STATE Compute the document reconstruction loss $L_{recons}$
		\STATE Compute the topic embedding regularization loss $R_{word}$
		\STATE Compute the word embedding regularization loss $R_{topic}$
		\STATE Update the model's parameters $\boldsymbol{\alpha}$, $\boldsymbol{\rho}$, $W$ with gradient descending
		\STATE Update the variational parameters $\boldsymbol{v_{\mu}}$, $\boldsymbol{v_{\sigma}}$ with gradient descending
		\ENDFOR
	\end{algorithmic} 
\end{algorithm}

\section{Experiment}
In this part, a series of comparative and analysis experiments are conducted to evaluate our method's performance. All related codes are implemented based on PyTorch\cite{paszke2019pytorch}, released after publication.

\subsection{Datasets}
We conduct experiments on two mainstream datasets, i.e., 20NewsGroup and New York Times(NYT), which are also used by nearly all neural topic models. For New York Times, we remove words whose document frequency is less than a certain number and obtain several sub-datasets with different vocabulary sizes. Table~\ref{tab:data} shows the statistical information of datasets we used. 
\begin{table}[!htb]
    \centering
    
    \scalebox{1.0}{
    \begin{tabular}[width=\linewidth]{|c|c|c|c|}
        \hline
        DATASET & Partition & \ Document Count & \ Vocabulary Size \\
        \hline
        \multirow{3}{*}{20NewsGroup} & train & 10132 &  \\
        \cline{2-3}
        & valid & 1126 & 1994 \\
        \cline{2-3}
        & test & 7487 & \\
        \hline
        \multirow{3}{*}{NYT-10000} & train & 	254616 &  \\
        \cline{2-3}
        & valid & 14978 & 1483 \\
        \cline{2-3}
        & test & 29934 &\\
        \hline
        \multirow{3}{*}{NYT-5000} & train & 254666 &  \\
        \cline{2-3}
        & valid & 14982 & 2889 \\
        \cline{2-3}
        & test & 29947 &\\
        \hline
        \multirow{3}{*}{NYT-3000} & train & 254671 &  \\
        \cline{2-3}
        & valid & 14982 & 4324 \\
        \cline{2-3}
        & test & 29952 &\\
        \hline
    \end{tabular}
    }
    \vspace{0.5em}
    \caption{Statistical information and data partition of datasets used in this paper. NYT-x, indicates removing words whose document frequency is less than x}
    \label{tab:data}
\end{table}{}
\subsection{Baselines}
We choose the following neural topic models as baselines to compare with our model.
\textbf{GSM} parameterizes the topic distribution as the Gaussian Softmax distribution, enabling both the inference and generative model to be jointly trained with back-propagation \cite{miao2017discovering}. 

\textbf{NTM} incorporates topic coherence as an additional training objective apart from the perplexity optimization target. Thus, NTM could achieve comparable perplexity and higher topic coherence. \cite{ding-etal-2018-coherence}

\textbf{NTMR} furthermore introduces a differentiable surrogate topic coherence regularization term on NTM, and the regularization term is measured as the weighted sum of word-to-topic cosine similarities \cite{ding-etal-2018-coherence}. 

\textbf{ETM} is a document model for large vocabulary and long-tailed language data. It models each word with a categorical distribution parameterized by the inner product of the word embedding to its assigned topic embedding \cite{dieng2020topic}. This model has a similar framework with our model, which leads to a similar performance to our model.

\subsection{Settings}
Our model is implemented by PyTorch 1.7.1 (cuda version) and all models are trained and tested on a 2080ti GPU.
To make our model easily reproduced, we display the hyper-parameters of our model on four datasets.
\begin{table}[!htb]
    \centering
    \begin{tabular}{|c|c|c|c|c|}
        \hline
        \multirow{2}{*}{DATASET} & LeakyRelu & \ Learning & \ Dropout & \ Embedding \\
        & Rate & Rate & Rate & Size\\
        \hline
        \hline
        20NewsGroup & 0.2 & 0.005 & 0.5 & 300\\
        \hline
        NYT-10000 & 0.2 & 0.005 & 0.5 & 300\\
        NYT-5000 & 0.2 & 0.009 & 0.5  & 300\\
        NYT-3000 & 0.2 & 0.006 & 0.5 & 300 \\
        \hline
    \end{tabular}
    \vspace{0.5em}
    \caption{Hyper-parameters of our model on different datasets.}
    \label{tab:param}
\end{table}{}

\subsection{Metrics}
To evaluate a topic model, we need to check the quality of generated topic-word distribution and the performance of document modeling. The former could be evaluated by topic coherence and topic diversity. The latter could be evaluated by perplexity. Next, we will introduce the metrics concretely.

\subsubsection{Evaluation for Topic Quality}
Following the ETM, we use the product of topic coherence and topic diversity as the overall performance for the quality of topics produced by the topic model:
\begin{equation}
    Quality = Coherence * Diversity\,.
\end{equation}

\textbf{Topic Coherence}
The calculation of topic coherence in this paper is the same as the setting in the ETM, which could be represented as the following:
\begin{equation}
    TC = \frac{1}{K}\sum_{k=1}^{K}\frac{1}{45}\sum_{i=1}^{10}\sum_{j=i+1}^{10}f(w_i^{(k)}, w_j^{(k)})\,,
\end{equation}
where $\{w_{1}^{(k)}, w_{2}^{(k)}, ... , w_{10}^{(k)}\}$ indicates top-10 most likely words in each topic $k$, $K$ is the number of topic, $f$ is the the normalized pointwise mutual information.
\begin{equation}
    f(w_i, w_j) = \frac{log\frac{p(w_i, w_j)}{p(w_i)p(w_j)}}{-logp(w_i, w_j)}\,,
\end{equation}
where, $p(w_i, w_j)$ is the probability of words
$w_i$ and $w_j$ co-occurring in a document and $p(w_i)$ is the probability of word $w_i$ in this corpus. In practice, we use empirical counts to replace the probability. In ETM's codes, $p(w_i)$ is represented by the number of documents containing word $w_i$ and $p(w_i, w_j)$ is represented by the number of documents contain word $w_i$ and word $w_j$ at the same time. A higher topic coherence is better.

\textbf{Topic Diversity}
Topic diversity we use is the same as ETM's topic diversity, which is the percentage of unique words in the top 25 words of all topics. When diversity is higher, topics are less redundant. Better topics should have a higher diversity.

\subsubsection{Evaluation for Document Modeling}
Neural topic models could output the probability of each word appearing in a document. In this way, we can compute each document's log-likelihood $logp(w_1, w_2, ..., w_N)$ where $w_i$ is the i-th word in this document. Also, a document's perplexity is calculated by:
\begin{equation}
    perplexity(d) = p(w_1,w_2,...,w_N)^{-\frac{1}{N}}\,,
\end{equation}
\begin{equation}
    \log(perplexity(d)) = -\frac{1}{N} logp(w_1,w_2,...,w_N)\,.
\end{equation}
Therefore, for a document set, its perplexity could represented as:
\begin{equation}
    perplexity = e^{L_{recons}}\,,
\end{equation}
where $L_{recos}$ is the topic model's reconstruction loss. In this paper, we use the average perplexity of each document in the test dataset as the final metric.

\subsection{Main Results}
\subsubsection{Results on 20NewsGroup}
\linespread{1.25}
\begin{table*}[!ht]
    \centering
    \resizebox{\textwidth}{20mm}{
    \begin{tabular}{|c||cccc|cccc|cccc|}
    \hline
    \multirow{3}{*}{METHODS} & \multicolumn{12}{c|}{Topic Number} \\
    \cline{2-13} 
    & \multicolumn{4}{c|}{K=50} & \multicolumn{4}{c|}{K=100} & \multicolumn{4}{c|}{K=200}   \\
    \cline{2-13}
	& Coherence &	Diversity &  Quality & Perplexity & Coherence &	Diversity &  Quality & Perplexity  &  Coherence &	Diversity &  Quality & Perplexity  	\\

\hline
NTMR & 0.1035 & 0.0776 & 0.0080 & 1135.1 & 0.1123 & 0.0472 & 0.0053 & 1060.7 & 0.1324 & 0.0126 & 0.0017 & 1174.8    \\
\hline
NTM	&	0.1804 & 0.3648	&	0.0658 & 897.4	&	0.1621 & 0.2592	&	0.0420 & 852.7	& 0.1037 & 0.1754	&	0.0182 & 823.8 \\
\hline
GSM	&	0.1605  & 0.3880	& 0.0623 & 828.0	&	0.1716 & 0.2824	&	0.0485 & 812.3	& 0.1524 & 0.1932	&	0.0294 & 785.5		 \\
\hline
ETM	&	0.1865 & 0.4864	&	0.0907 & 686.0	&	0.1821 & 0.3552	&	0.0647 & 660.0	&	0.1826 &  0.2326	&	0.0425 & 681.0\\
\hline
ERNTM(ours)	&	\textbf{0.1949} & \textbf{0.5112}	& \textbf{0.0996} & \textbf{651.1}	&	\textbf{0.1873} & \textbf{0.3624}	&	\textbf{0.0679} & \textbf{653.9}	&	\textbf{0.1867} & \textbf{0.2360}	&	\textbf{0.0441} & \textbf{671.2}      \\
    \hline
    \end{tabular}}
    \linespread{1}
    \caption{Comparison with baselines on 20NewsGroup datasets. Here we explore the models' performance on a different number of topics. K means the number of topics. All experiments use the same random seed. The best value is bolded.}
    \label{tab:results_20}
\end{table*}

Here, we adopted the same hyper-parameters as NTM's codes to compare fairly. Also, we found that using a pre-trained word embedding instead of initializing it randomly in some neural topic models, such as NTM, NTMR, will lead to worse results. To compare with the best baselines results, we use randomly initialized word embedding and topic embedding in all experiments. Besides, we use the same random seed (2020) in all models to overcome the influence of different initial parameters. Experiments on other datasets also follow this setting.

Table~\ref{tab:results_20} shows the comparison results on the 20NewsGroup dataset with different topic numbers. According to this table, our model achieves the best performance on topic quality compared with other baseline models when set with various topic numbers. In detail, for performance on topic diversity and topic coherence, our model both outperforms other models significantly. Besides, our model is still the best for the performance on document modeling evaluated by perplexity. 

However, we can find that with the increasing number of topics, the gain of our model compared with ETM gradually diminishes. A possible reason is that more topics mean more points in topic embedding space. According to the sampling theory in the high dimension space, many topic embeddings are already orthogonal, which is the aim of our constraint on the topic embedding. Therefore the role of our constraint on the topic embedding is weakened.


\subsubsection{Results on New York Times}
\linespread{1.25}
\begin{table*}[!ht]
    \centering
    \resizebox{\textwidth}{20mm}{
    \begin{tabular}{|c||cccc|cccc|cccc|}
    \hline
    \multirow{3}{*}{METHODS} & \multicolumn{12}{c|}{Datasets} \\
    \cline{2-13} 
    & \multicolumn{4}{c|}{NYT-10000} & \multicolumn{4}{c|}{NYT-5000} & \multicolumn{4}{c|}{NYT-3000}   \\
    \cline{2-13}
	& Coherence &	Diversity &  Quality & Perplexity & Coherence &	Diversity &  Quality & Perplexity  &  Coherence &	Diversity &  Quality & Perplexity  	\\

\hline
NTMR & 0.078 & 0.3952 & 0.0307 & 881.4 & 0.0711 & 0.4224 & 0.0300 & 1358.3 & 0.0929 & 0.5504 & 0.0511 & 1734.0    \\
\hline
NTM	&	0.1811 & 0.4200	&	0.0761 & 679.0	&	0.1924 & 0.5552	&	0.1068 & 1066.1	& 0.2011 & 0.6064	&	0.1219 & 1377.7 \\
\hline
GSM	&	\textbf{0.2410}  & 0.4176	& 0.1006 & 775.3	&	\textbf{0.2847} & 0.4840	&	0.1378 & 1293.0	& \textbf{0.3047} & 0.5168	&	\textbf{0.1575} & 1699.5		 \\
\hline
ETM	&	0.1885 & 0.6224	& 0.1173 & \textbf{642.1}	&	0.2003 & 0.6416	&	0.1285 & 1064.7	&	0.2083 & 0.6704	&	0.1397 & 1372.7      \\
\hline
ERNTM(ours)	&	0.1888 & \textbf{0.6256}	&	\textbf{0.1181} & 644.1	&	0.2104 & \textbf{0.6768}	&	\textbf{0.1424} & \textbf{1060.2}	&	0.2157 &  \textbf{0.7096}	&	0.1531 & \textbf{1365.9}\\

    \hline
    \end{tabular}}
    \linespread{1}
    \caption{Comparison with baselines on New York Times datasets. Here we explore the models' performance on different vocabulary sizes. The topic number is set to 50. All experiments use the same random seed. The best value is bolded.}
    \label{tab:results_nyt}
\end{table*}
Table~\ref{tab:results_nyt} shows the comparison results on three New York Times datasets with different vocabulary sizes. The topic number is set to 50. We still use randomly initialized word embedding and topic embedding in all experiments on these datasets.

According to this table, our model has the best topic quality on NYT-10000 and NYT-5000 datasets. Although GSM outperforms our model marginally for topic quality on the NYT-3000 dataset, our model is significantly superior to GSM on the perplexity. On NYT-5000 and NYT-3000 datasets, our model has the lowest perplexity. For the perplexity on the NYT-10000 dataset, ETM has a better result than our model. However, its performance on topic quality is lower than our model. GSM has the best performance on topic coherence and achieves the best topic quality value on the NYT-3000 dataset, but it has a significantly higher perplexity than most other models. And our model's topic quality value on the NYT-3000 dataset is comparable with GSM's topic quality value. For topic diversity, our model is the best one on all datasets. Generally, our model exhibits better adaptability than other baselines on datasets with different vocabulary sizes.

There are also some interesting results in this table. Our model shows a better document modeling performance on datasets with a larger vocabulary. A possible reason is that a larger vocabulary means that more information flows into word embedding, which is helpful to document modeling. Compared with ETM, our model's gain on topic quality also increases with the vocabulary size, which shows the effect of constraints on embeddings.

\subsection{Influence of Topic Number and Vocabulary Size}
The topic number and vocabulary size are two critical hyper-parameters to topic models. The results in Table~\ref{tab:results_20} and Table~\ref{tab:results_nyt} show how the number of topics and vocabulary size influences the neural topic model's performance. In detail, according to the results on the 20NewsGroup dataset, the topic quality of neural topic models is lower with the increasing topic number. This trend is evident for all models. Therefore, we could assume that more topic numbers will worsen topic quality. However, the performance change on perplexity does not show a consistent trend. A possible explanation is that topic number is not the most pivotal factor to perplexity. 

In Table~\ref{tab:results_nyt}, results on datasets with different vocabulary size is presented. NYT-10000 has the least vocabulary size, and NYT-3000 has the largest vocabulary. According to this table, we can observe two obvious trends. The first trend is that with the increasing vocabulary size, the topic quality of neural topic models will be improved. Another trend is that a larger vocabulary will lead to larger perplexity, which means worse performance on document modeling. The above trends indicate that vocabulary size could dominantly influence the topic quality and document modeling performance of topic models.

\subsection{Ablation Study on Model Performance}

To further explore the effects of two constraints on topic embedding and word embedding, we performed an ablation study on 20NewsGroup dataset with the topic number set to $200$. The results are shown in Table~\ref{tab:ablation}. According to this table, when only word embedding constraint is utilized, the results are the worst on all metrics. When we use both constraints on topic and word embedding, the results are the best on all metrics. Therefore, we can know that only applying the word embedding constraint is useless and even harmful to the model's performance. While we use two constraints simultaneously, there is an extra benefit to model performance. When only the constraint on topic embedding works, the topic diversity of the model is improved. In summary, we can find that the constraint on topic embedding is more influential than the constraint on word embedding. Also, when two constraints are combined, the performance will be improved significantly.
\begin{table}[!htb]
\small
    \centering
    \begin{tabular}{|c|c|c|c|c|c|}
    \toprule
    Topic & Word & Coherence & Diversity & Quality & Perplexity \\
    \midrule
     - & - &  0.1832 &  0.2266 & 0.0415 & 672.3 \\
     - & + &  0.1814 &  0.2180 & 0.0395 & 680.2 \\
     + & - & 0.1827 & 0.2284 & 0.0417 & 671.5 \\
     + & + & \textbf{0.1867} & \textbf{0.236} & \textbf{0.0441} & \textbf{671.2}  \\
    \bottomrule
    \end{tabular}
    \vspace{0.5em}
    \caption{Results of ablation study on the 20NewsGroup dataset. The best value is bolded}
    \label{tab:ablation}
\end{table}{}

\subsection{Ablation Study based on Uniformity}
Uniformity is proposed by \cite{Wang2020UnderstandingCR} and used to evaluate the quality of the representation space. The calculation is shown as:
\begin{equation}
    L_{uniformity} = \mathop{E}_{(x, y)\sim{p_{data}}}e^{-2||f(x) - f(y)||^2},
\end{equation}
where $(x, y)$ is the pair consisting of any two different representations, $f(x)$ is the normalized representation of $x$. A lower uniformity means a well-distributed representation space. we hope to display the changes in these embeddings during training quantitatively. So, when they achieve better performance on the valid dataset, we calculate the uniformity value of the topic embedding, word embedding and topic-word distribution produced by our original model and its variants (remove one or two constraints). After obtaining a value list for each model, we choose points on the condition that at least 50 training epochs separate two adjacent points. The uniformity changes of topic embedding, word embedding and topic-word distribution are present by Fig.~\ref{fig:topic_emb}, Fig.~\ref{fig:word_emb}, and Fig.~\ref{fig:topic_word}.

For the topic embedding, according to the Fig.~\ref{fig:topic_emb}, only removing the loss term of word embedding nearly makes no influence to the uniformity of the topic embedding, which means the topic embedding could not obtain any information from the loss term in the training process. While we remove the loss term of topic embedding, the uniformity increases, which means the topic representation space is worse. So the constraint on topic embedding could help produce a better representation space of the topic. Besides, all models' uniformity decrease with the training process. On this basis, we could assume that a topic embedding space with low uniformity is expected in the embedding-based neural topic model.

\begin{figure}
    \centering
    \includegraphics[width=\linewidth]{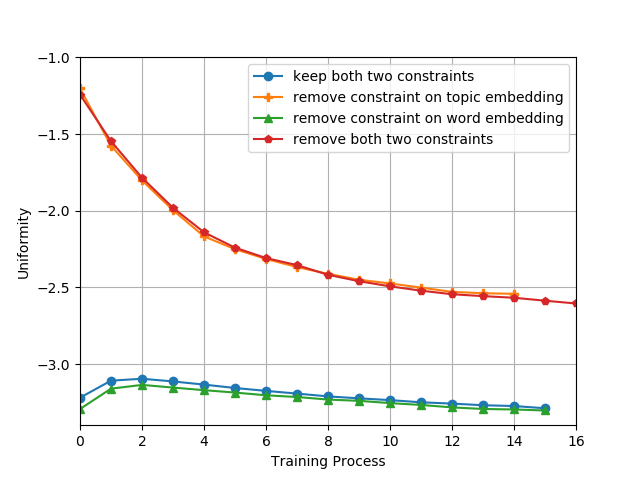}
    \caption{The change of the topic embedding's uniformity in the training process. Y axis represents uniformity value and X axis means epoch of training process (Each point represents 50 epochs).}
    \label{fig:topic_emb}
\end{figure}

The Fig.~\ref{fig:word_emb} shows the change of the word embedding's uniformity. Unlike topic embedding, removing the loss term of topic embedding leads to a slight increase in the uniformity of the word embedding. Also, when we keep the loss term of word embedding in the training process, the uniformity increases, which means the topic representations are worse. This phenomenon also implies damage from the constraint on word embedding to the uniformity, consistent with the ablation study on model performance results. A possible reason is that the representations of documents are spare, which leads to more words with close embedding representations. This result means a higher uniformity value. Also, we could find that the uniformity of the word embedding has a limited influence on the model performance.

\begin{figure}
    \centering
    \includegraphics[width=\linewidth]{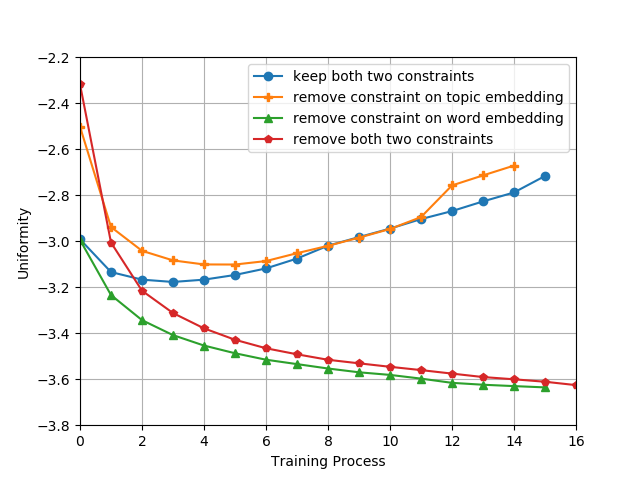}
    \caption{The change of the word embedding's uniformity in the training process. Y axis represents uniformity value and X axis means epoch of training process (Each point represents 50 epochs).}
    \label{fig:word_emb}
\end{figure}

The Fig.~\ref{fig:topic_word} displays the change of the word embedding's uniformity. We can find that the model with two constraints has the best uniformity value and the model only utilizing constraint of the word embedding achieves the worst uniformity. It is surprisingly consistent with the result in the ablation study on model performance. Also, similar to topic embedding and word embedding, the uniformity of topic-word distribution decreases gradually in the training process. The result shows that a topic-word distribution with low uniformity is expected in the embedding-based neural topic model. And low uniformity means higher topic diversity. 

\begin{figure}
    \centering
    \includegraphics[width=\linewidth]{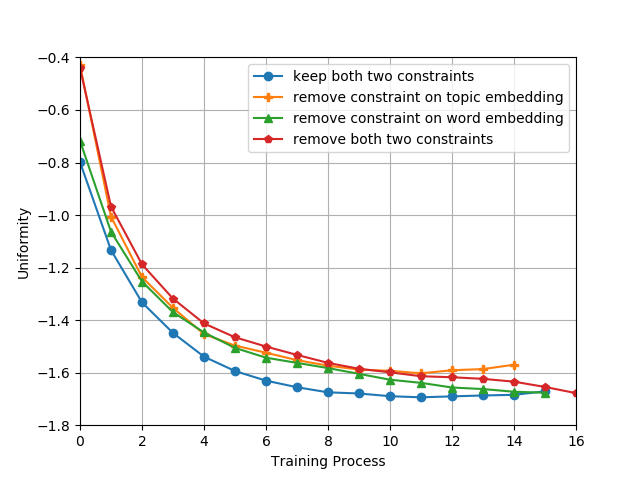}
    \caption{The change of the topic-word distribution's uniformity in the training process. Y axis represents uniformity value and X axis means epoch of training process (Each point represents 50 epochs).}
    \label{fig:topic_word}
\end{figure}

\section{Conclusion}

In this paper, we propose an embedding regularized neural topic model to explicitly constrain the training of embeddings in the embedding-based neural topic model. Also, we firstly introduce uniformity to describe the changes in embeddings and thus give a detailed exploration of how embeddings work in the embedding-based neural topic model via ablation study. A series of experiments demonstrated our model's competitiveness over previous methods. Our contribution to the interpretability of the embedding-based neural topic model reveals important intuition in designing future embedding-based neural topic models. In the future, we will explore which NLP problems could obtain benefits from our method.

\section*{Acknowledgment}

This work was supported in part by the Hong Kong RGC grant ECS 21212419, the Technological Breakthrough Project of Science, Technology and Innovation Commission of Shenzhen Municipality under Grants JSGG20201102162000001, InnoHK initiative, the Government of the HKSAR, Laboratory for AI-Powered Financial Technologies, the Hong Kong UGC Special Virtual Teaching and Learning (VTL) Grant 6430300, and the Tencent AI Lab Rhino-Bird Gift Fund.
\bibliographystyle{IEEEtran}
\bibliography{ref}
\end{document}